\documentclass[11pt,a4paper]{article}
\usepackage{times,latexsym}
\usepackage{url}
\usepackage[T1]{fontenc}
\usepackage{amsmath}
\usepackage{amssymb}
\usepackage{booktabs}
\usepackage{multicol}
\usepackage{multirow}
\usepackage{graphicx}
\usepackage{subfigure}

\usepackage[acceptedWithA]{tacl2018v2}

\usepackage{xspace,mfirstuc,tabulary}

\newif\iftaclinstructions
\taclinstructionsfalse % AUTHORS: do NOT set this to true
\iftaclinstructions
\renewcommand{\confidential}{}
\renewcommand{\anonsubtext}{(No author info supplied here, for consistency with
TACL-submission anonymization requirements)}
\newcommand{\instr}
\fi

\iftaclpubformat % this "if" is set by the choice of options

\else

\fi

\def\tinycol{\hskip 4pt}

\title{How Can We Know \emph{When} Language Models Know? \\ On the Calibration of Language Models for Question Answering}

\author{
Zhengbao Jiang$^\dag$, Jun Araki$^\ddag$, Haibo Ding$^\ddag$, Graham Neubig$^\dag$ \\
  $^\dag$Languages Technologies Institute, Carnegie Mellon University \\
  $^\ddag$Bosch Research \\
\texttt{\{zhengbaj,gneubig\}@cs.cmu.edu} \\
\texttt{\{jun.araki,haibo.ding\}@us.bosch.com}
}
\date{}

\begin{document}
\maketitle
\begin{abstract}
Recent works have shown that language models (LM) capture different types of knowledge regarding facts or common sense.
However, because no model is perfect, they still fail to provide appropriate answers in many cases.
In this paper, we ask the question ``how can we know when language models know, with confidence, the answer to a particular query?''
We examine this question from the point of view of \emph{calibration}, the property of a probabilistic model's predicted probabilities actually being well correlated with the probabilities of correctness.
We examine three strong generative models -- T5, BART, and GPT-2 -- and study whether their probabilities on QA tasks are well calibrated, finding the answer is a relatively emphatic \emph{no}.
We then examine methods to calibrate such models to make their confidence scores correlate better with the likelihood of correctness through fine-tuning, post-hoc probability modification, or adjustment of the predicted outputs or inputs.
Experiments on a diverse range of datasets demonstrate the effectiveness of our methods.
We also perform analysis to study the strengths and limitations of these methods, shedding light on further improvements that may be made in methods for calibrating LMs.
We have released the code at \url{https://github.com/jzbjyb/lm-calibration}.
\end{abstract}

\section{Introduction}

\begin{table*}[t]
\small
\centering
\begin{tabular}{@{}l@{\tinycol}p{0.33\textwidth}|p{0.30\textwidth}@{\tinycol}r@{\tinycol}r@{}}
\toprule
\textbf{Format} & \textbf{Input} & \textbf{Candidate Answers} & \textbf{Original} & \textbf{Calibrated} \\
\midrule
\multirow{4}{*}{Multiple-choice} & \multirow{4}{*}{\parbox{0.33\textwidth}{Oxygen and sugar are the products of (A) cell division. (B) digestion. (C) photosynthesis. (D) respiration.}} & cell division. & 0.00 & 0.02 \\
 & & digestion. & 0.00 & 0.01 \\
 & & \textbf{photosynthesis.} & 0.00 & 0.83 \\
 & & respiration. & 1.00 & 0.14 \\
\midrule
\multirow{4}{*}{Extractive} & \multirow{4}{*}{\parbox{0.33\textwidth}{What type of person can not be attributed civil disobedience? \\ Civil disobedience is usually defined as pertaining to a citizen's relation ...}} & \textbf{head of government} & 0.07 & 0.49 \\
 & & public official & 0.91 & 0.26 \\
 & & head of government of a country & 0.01 & 0.16 \\
 & & public officials & 0.01 & 0.09 \\
\bottomrule
\end{tabular}
\caption{LM calibration examples for the T5 model with correct answers in bold. ``Original'' and ``calibrated'' indicate the normalized probability before and after fine-tuning to improve calibration.}
\label{tab:example}
\end{table*}

Language models (LMs; \citet{church-1988-stochastic,bengio2003neural,radford-2019-gpt2}) learn to model the probability distribution of text, and in doing so capture information about various aspects of the syntax or semantics of the language at hand.
Recent works have presented intriguing results demonstrating that modern large-scale LMs also capture a significant amount of knowledge, including factual knowledge about real-world entities \cite{petroni-etal-2019-language,jiang-2019-lpaqa,roberts-2020-pack,bouraoui-2020-relbert}, commonsense knowledge \cite{trinh-2018-commonsense,kocijan-2019-wsc,talmor-2019-olmpics,bosselut-2019-comet}, and simple numerical operations \cite{wallace-2019-numemb,talmor-2019-olmpics,geva-2020-genbert}.
Notably, large models trained on massive crawls of internet text (such as T5 \cite{raffel-2019-t5} and GPT-3 \cite{brown-2020-gpt3}) have been shown to be able to perform quite sophisticated knowledge-based tasks simply through prompting the model to predict the next words given a particular cue.

However, at the same time, LMs are obviously not omnipotent, and still fail to provide appropriate answers in many cases, such as when dealing with uncommon facts \cite{poerner-2019-ebert,jiang-2020-xfactr} or complex reasoning \cite{talmor-2019-olmpics}.
The high performance on datasets probing factual or numerical knowledge might be achieved through modeling superficial signals in the training data that are not generalizable to unseen test cases \cite{poerner-2019-ebert,zhou-2020-bertinfer,wallace-2019-numemb,talmor-2019-olmpics}.
Thus, if such models are to be deployed in real applications it is of crucial importance to determine the \emph{confidence} with which they can provide an answer.
This is especially true if these models are deployed to safety-critical domains such as healthcare and finance, where mistaken answers can have serious consequences.%
\footnote{For example, a mocked-up medical chatbot based on GPT-3 answered the question of ``should I kill myself?'' with ``I think you should'' \cite{quach20gpt3suicide}.}

In this paper, we ask the question ``how can we know when language models know, with confidence, the answer to a particular knowledge-based query?''
Specifically, we examine this from the point of view of \emph{calibration}, whether the model's probability estimates are well-aligned with the actual probability of the answer being correct.
We apply the largest publicly available LMs, T5, BART, and GPT-2, over a wide range of question answering (QA) datasets \cite{khashabi-2020-unifiedqa} covering diverse domains.
We first observe that despite the models' high performance (e.g.~T5 eclipses other alternatives such as GPT-3 on some datasets), the models tend to not be well calibrated; their probability estimates over candidates have far-from-perfect correspondence with the actual probability that the answer they provide is correct.
Some examples of this are demonstrated in the ``Original'' column of \autoref{tab:example}.

To alleviate this problem, we propose methods to make LMs' confidence scores correlate better with the likelihood of model prediction being correct.
We examined both fine-tuning methods that modify LMs' parameters and post-hoc methods that keep LMs fixed and only manipulate the confidence values or inputs.
Specifically, we fine-tune the LM using softmax- or margin-based objective functions based on multiple candidate answers.
For post-hoc calibration, we examined temperature-based scaling and feature-based decision trees that take prediction probability and input-related features as input and produce calibrated confidence \cite{jagannatha-2020-structcal,desai-2020-transcal,kamath-2020-qacal}.
We also study the sensitivity of LMs' confidence estimation with respect to language variation by paraphrasing candidate answers and augmenting questions using retrieved context.

Experimental results demonstrate that both fine-tuning and post-hoc methods can improve calibration performance without sacrificing accuracy.
We further perform analysis and ablation studies on our methods, inspecting different aspects that may affect calibration performance.
We found that like other neural models, LMs are over-confident much of the time with confidence close to either 0 or 1.
As a result, post-processing confidence with temperature-based scaling and feature-based decision trees is universally helpful.
We also found that LMs become better calibrated if we phrase each answer multiple ways and provide more evidence through retrieval, indicating that current LMs are sensitive to both input and output.

\section{LM-based Question Answering}

LMs are now a ubiquitous tool in not only natural language generation, but also natural language understanding (NLU), where they are largely used for unsupervised representation learning in pre-trained models such as BERT \citep{devlin-etal-2019-bert}.
However, recent work has demonstrated that LMs can also be used \emph{as-is} to solve NLU tasks, by predicting the missing words in Cloze-style questions \citep{petroni-etal-2019-language}, or by predicting the continuation to prompts \cite{bosselut-2019-comet,brown-2020-gpt3}.

Previous works that purport to calibrate LMs \cite{desai-2020-transcal,jagannatha-2020-structcal,kamath-2020-qacal,kong-2020-lminout} mainly focus on the former use case, using representations learned by LMs to predict target classes (for tasks such as natural language inference, part-of-speech tagging, or text classification) or identify answer spans (for tasks such as extractive QA).
In contrast, we focus on the latter case, calibrating LMs themselves by treating them as natural language generators that predict the next words given a particular input.

To make our observations and conclusions as general as possible, we experiment over a diverse range of QA datasets with broad domain coverage over questions regarding both factual and commonsense knowledge \cite{khashabi-2020-unifiedqa}.
We list all the datasets we used in \autoref{tab:dataset} along with their corresponding domain.
Since we focus on calibrating LMs as generators, we follow \citet{khashabi-2020-unifiedqa} in converting QA datasets of different formats to a unified sequence-to-sequence format that takes a question $X$ as input and calculates the probability of a continuation $Y$ that corresponds to the answer:
\begin{equation*}
P_{\text{LM}}(Y|X) = \prod_{i=1}^{|Y|} P_{\text{LM}}(y_i|X, y_{<i}).
\end{equation*}

Specifically, we focus on two varieties of QA: \emph{multiple-choice} and \emph{extractive}, with examples shown in \autoref{tab:example}.%
\footnote{We also considered using free-form (abstractive) QA datasets, where the answers are not constrained to be one of several choices and can instead be any text. However, we found it hard to evaluate the correctness of generated outputs, as paraphrases of the correct answer are still correct, so we do not report results on these datasets in this paper. Solving this evaluation problem and evaluating calibration on these tasks is an enticing direction for future work.}

\paragraph{Multiple-choice QA}
For multiple-choice QA, we assume a question and a set of candidate answers $\mathcal{I}(X)=\{Y^{(i)}\}_i$.
Inputs $X$ to LMs are questions concatenated with multiple candidate answers (with each answer prefaced by ``(A)'', ``(B)'', etc.), and context such as a passage that can be used to help answer the question if any exists.
To find the answer the model will return, we calculate the highest-probability answer among the answer candidates:
\begin{equation*}
    \hat{Y} = \underset{Y' \in \mathcal{I}(X)}{\arg\max} P_{\text{LM}}(Y'|X).
\end{equation*}
We can also calculate the normalized probability
\begin{equation}
    P_{N}(\hat{Y}|X) = \frac{P_{\text{LM}}(\hat{Y}|X)}{\sum_{Y' \in \mathcal{I}(X)}P_{\text{LM}}(Y'|X)},\label{eq:norm}
\end{equation}
which provides some idea of the confidence of answer $\hat{Y}$ with respect to the candidate list.

\paragraph{Extractive QA}
For extractive QA, inputs $X$ to LMs are questions concatenated with context passages from which the answer must be extracted.
In this case, every span within the passage is a candidate answer in $\mathcal{I}(X)$.
However, enumerating over all possible spans of the context passage is computationally costly.
Thus, we follow \citet{jagannatha-2020-structcal} in using a manageable set of candidate outputs to perform calibration.
Specifically, we develop a method to efficiently calculate probabilities over promising spans that exist in the input.
First, we calculate the probability of the first token in output $Y'$, masking out any tokens that are not included in the input passage at all.
Then, for the top $R$ scoring tokens, we find their location in the input passage, and calculate the probability of all continuing spans up to a certain length (e.g., 20 tokens).
We finally keep the top $K$ spans as candidates $\mathcal{I}(X)$ and use all candidates to calculate the probability in a manner similar to that of multiple-choice QA.

\begin{table}[t]
\small
\centering
\begin{tabular}{@{}l|p{0.7\columnwidth}@{}}
\toprule
\textbf{Format} & \textbf{Datasets and Domains} \\
\midrule
Multi-choice & ARC (science \cite{clark-2018-arc}), AI2 Science Questions (science \cite{clark-2018-arc}), OpenbookQA (science \cite{mihaylov-2018-openbookqa}), Winogrande (commonsense \cite{sakaguchi-2020-wino}), CommonsenseQA (commonsense \cite{talmor-2019-commonsenseqa}), MCTest (fictional stories \cite{richardson-2013-mctest}), PIQA (physical \cite{bisk-2020-piqa}), SIQA (social \cite{sap-2019-siqa}), RACE (English comprehension \cite{lai-2017-race}), QASC (science \cite{khot-2020-qasc}), MT-test (mixed \cite{hendrycks-2020-mass}) \\
Extractive & SQuAD 1.1 (wikipedia \cite{rajpurkar-2016-squad}), SQuAD 2 (Wikipedia \cite{rajpurkar-2018-squad2}), NewsQA (news \cite{trischler-2017-newsqa}), Quoref (wikipedia \cite{dasigi-2019-quoref}), ROPES (situation understanding \cite{lin-2019-ropes}) \\
\bottomrule
\end{tabular}
\caption{Datasets used in this paper and their domains.}
\label{tab:dataset}
\end{table}

\section{Background on Calibration}

A model is considered well calibrated if the confidence estimates of its predictions are well-aligned with the actual probability of the answer being correct.
Given an input $X$ and true output $Y$, a model output $\hat{Y}$, and a probability $P_{N}(\hat{Y}|X)$ calculated over this output, a perfectly calibrated model satisfies the following condition:
\begin{equation*}
P(\hat{Y}=Y|P_N(\hat{Y}|X)=p) = p, \forall p \in [0,1].
\end{equation*}
In practice, we approximate this probability by bucketing predictions into $M$ disjoint equally-sized interval bins based on confidence.
\citet{guo-2017-cal} examined the calibration properties of neural network classifiers, and proposed a widely used measure of calibration called expected calibration error (ECE), which is a weighted average of the discrepancy between each bucket's accuracy and confidence:
\begin{equation}
\sum_{m=1}^{M}{\frac{|B_m|}{n}|\text{acc}(B_m) - \text{conf}(B_m)|},
\label{eq:ece}
\end{equation}
where $B_m$ is the $m$-th bucket containing samples whose prediction confidence falls into the interval $(\frac{m-1}{M}, \frac{m}{M}]$, $\text{acc}(B_m)$ is the average accuracy of this bucket, and $\text{conf}(B_m)$ is the average confidence of this bucket.
The above equation can be visualized using reliability diagrams (e.g., \autoref{fig:reliability} in the experiments), where each bar corresponds to one bucket, and the height is equal to the average accuracy.
The diagram of a perfectly calibrated model should have all bars aligned with the diagonal.

Unfortunately, we found that state-of-the-art LM-based methods for question answering (such as the UnifiedQA model of \citet{khashabi-2020-unifiedqa}) were extraordinarily poorly calibrated, with the normalized probability estimates barely being correlated with the likelihood of the outputs being correct.
For the two examples in \autoref{tab:example}, for instance, we can see that the language model assigns a very high probability to answers despite the fact that they are wrong.
This is particularly important because with T5 \citep{raffel-2019-t5}, GPT-3 \citep{brown-2020-gpt3}, and others \citep{guu-2020-realm,lewis-2020-rag} being provided as a potential answer to complex knowledge-based tasks, for models to actually be used in practical scenarios they must also be able to know when they cannot provide correct information.
In the following section, we examine methods to improve the calibration of pre-trained models through a number of methods.

\section{Calibrating LMs for Question Answering}
Our calibration methods can be grouped into two categories: methods that fine-tune LMs and post-hoc methods that keep LMs fixed and only manipulate confidence or inputs.

\subsection{Fine-tuning-based Calibration}
Existing LMs mainly use maximal likelihood estimation (MLE) during training, which maximizes the probability of ground truth output given the input.
However, it is well-attested that MLE-trained language generators are biased, tending to prefer short outputs \citep{murray-chiang-2018-correcting}, or being biased towards more frequent vocabulary \citep{ott2018analyzing}.
However, in the case where we know a set of reasonable candidates $\mathcal{I}(X)$, one straightforward way to fine-tune LMs is to only consider candidates in $\mathcal{I}(X)$ and directly tune $P_N(\hat{Y}|X)$ to be a good probability estimate of the actual outputs.
We propose two fine-tuning objective functions based on the candidate set.

\paragraph{Softmax-based} objective functions model candidates in a one-vs-all setting, where we use the softmax function to normalize the confidence of candidates and maximize the probability corresponding to the correct candidate.
We use the negative log likelihood as the loss function:
\begin{equation*}
L(X, Y) = - \log \frac{\exp(s(Y))}{\sum_{Y' \in \mathcal{I}(X)}{\exp(s(Y'))}},
\end{equation*}
where the ground truth $Y$ is one of the candidates in $\mathcal{I}(X)$, and $s(\cdot)$ is the logit of the corresponding output (omit condition $X$ for simplicity), which is computed as the log probabilities of all tokens in the output: $s(Y) = \log P_{\text{LM}}(Y|X)$.

\paragraph{Margin-based} objective functions try to maximize the confidence margin between ground truth output and negative results. This is motivated by the fact that non-probabilistic objectives such as those used by support vector machines provide reasonably good probabilistic estimates after appropriate scaling and adjustment \citep{platt1999probabilistic}. Specifically, we use the following objective:
\begin{equation*}
L(X, Y) = \sum_{Y' \in \mathcal{I}(X) \setminus Y}{\max(0, \tau + s(Y') - s(Y))}.
\end{equation*}

\subsection{Post-hoc Calibration}
Comparing to fine-tuning methods that optimize the parameters in the model, post-hoc calibration methods keep the model as-is and manipulate various types of information derived from the model to derive good probability estimates \cite{guo-2017-cal,jagannatha-2020-structcal,desai-2020-transcal}.
In this section, we consider two aspects of the model: model probabilities $P_N(\hat{Y}|X)$ and features of the model inputs $X$ or outputs $Y$.
We attempted two representative methods, namely temperature-based scaling \cite{guo-2017-cal} and feature-based decision trees \cite{jagannatha-2020-structcal}, to study whether post-processing probabilities is an effective method for calibration of LMs in the context of QA.

\paragraph{Temperature-based Scaling} methods have been proposed for classification tasks \cite{guo-2017-cal,desai-2020-transcal}, where a positive scalar temperature hyperparameter $\tau$ is introduced in the final classification layer to make the probability distribution either more peaky or smooth: $\text{softmax}(\mathbf{z} / \tau)$.
If $\tau$ is close to 0, the class with the largest logit receives most of the probability mass, while as $\tau$ approaches $\infty$, the probability distribution becomes uniform.
When applying this method to our setting, we use log probabilities of the candidates in $\mathcal{I}(X)$ as logits in computing the softmax function: $z = \log P_{\text{LM}}(Y'|X), Y' \in \mathcal{I}(X)$, and $\tau$ is optimized with respect to negative log likelihood on the development split.

\paragraph{Feature-based Decision Trees} methods explore non-linear combinations of features to estimate the confidence compared to temperature-based scaling which only considers the raw confidence.
We follow previous works \cite{jagannatha-2020-structcal,dong-2018-parsercal} and use gradient boosted decision trees \cite{chen-2016-xgb} as our regressor to estimate the confidence based on features.
Besides the raw confidence, we consider the following features and explain their intuitions:
\begin{itemize}
\item \textbf{Model Uncertainty}: We use the entropy of the distribution over the candidate set $\mathcal{I}(X)$ to inform the regressor of how uncertain the LM is with respect to the question.
\item \textbf{Input Uncertainty}: We use the perplexity of the LM on the input to indicate the uncertainty over the input. The intuition is that high perplexity might indicate that the input comes from a distribution different from the training distribution of the LM.
\item \textbf{Input Statistics}: We also use the length of the input and output as features, motivated by our hypothesis that longer text may provide more information to LMs than shorter text.
\end{itemize}
We train the regressor on the development set similarly to temperature-based scaling by minimizing negative log likelihood.

\subsection{LM-Specific Methods}\label{sec:method_lm}

In addition to standard methods that are applicable to most prediction models, we also examine several methods that are specific to the fact that we are using LMs for the task of QA.

\paragraph{Candidate Output Paraphrasing}

\begin{table}[t]
\small
\centering
\begin{tabular}{@{}p{0.27\columnwidth}|p{0.68\columnwidth}@{}}
\toprule
Input & How would you describe Addison? (A) excited (B) careless \textbf{(C) devoted}. Addison had been practicing for the driver's exam for months. He finally felt he was ready, so he signed up and took the test. \\
\midrule
Paraphrases \& Probabilities & devoted (0.04), dedicated (0.94), commitment (0.11), dedication (0.39) \\
\bottomrule
\end{tabular}
\caption{An example question with the correct answer in bold. Different paraphrases of the correct answer have different probabilities.}
\label{tab:para}
\end{table}

Motivated by the fact that LMs are sensitive to language variation \cite{jiang-2019-lpaqa} in tasks like question answering and factual prediction, we hypothesize that one potential reason why the confidence estimation of LMs is not accurate is that the candidate output is not worded in such a way that the LM would afford it high probability.
As shown by the example in \autoref{tab:para}, paraphrasing the correct answer from ``devoted'' to ``dedicated'' increases the probability from 0.04 to 0.94.
Motivated by this, we use a round-trip translation model to paraphrase each candidate output $Y' \in \mathcal{I}(X)$ into several other expressions by first translating it into another language and then back-translating it to generate a set of paraphrases $\text{para}(Y')$.
We then calculate the probability of each candidate output by summing the probability of all paraphrases $P(Y')=\sum_{Q \in \text{para}(Y')} P_{\text{LM}}(Q|X)$ and normalize it following \autoref{eq:norm}.
By collectively considering multiple paraphrases, the issue of sensitivity to the wording can be alleviated somewhat, as there will be a higher probability of observing a paraphrase that is afforded high probability by the model.

\paragraph{Input Augmentation}
Previous work has found that LMs' factual predictions can be improved if more context is provided \cite{petroni-2020-contextlm}, which has inspired many retrieval-augmented LMs that retrieve evidence from external resources and condition the LMs' prediction on this evidence \cite{guu-2020-realm,lewis-2020-marge,lewis-2020-rag}.
We hypothesize that retrieving extra evidence to augment the input also has the potential to improve the confidence estimation of LMs as it will provide the model more evidence upon which to base both its predictions and its confidence estimates.
We follow \cite{petroni-2020-contextlm} to retrieve the most relevant Wikipedia article using TF-IDF-based retrieval systems used in DrQA \cite{chen-2017-drqa} and append the first paragraph of the article to the input.

\section{Experiments}\label{sec:exp}
\subsection{Experimental Settings}
\paragraph{Datasets}
We evaluate the calibration performance on both multiple-choice QA datasets and extractive QA datasets listed in \autoref{tab:dataset}.
To test whether our calibration methods can generalize to out-of-domain datasets, we use a subset of datasets of multiple-choice/extractive QA to train our methods, and the remaining subset of datasets to evaluate the performance.
Specifically, we use ARC (easy), AI2 Science Question (elementary), OpenbookQA, QASC, Winogrande, CommonsenseQA, and PhysicalIQA as the training subset for multiple-choice QA (denoted as \textbf{MC-train}), and SQuAD 1.1, NewsQA as the training subset for extractive QA (denoted as \textbf{Ext-train}).
The remaining subsets used for evaluation are denoted as \textbf{MC-test} and \textbf{Ext-test} respectively.
We also included a much harder multiple-choice QA dataset (denoted as \textbf{MT-test}; \citet{hendrycks-2020-mass}) regarding common sense in a number of genres, in which the largest GPT-3 model and UnifiedQA both display only low to moderate accuracy.
For fine-tuning methods, we use the train split of MC-train/Ext-train to fine-tune the LMs.
For post-hoc methods like temperature-based scaling and decision trees, we follow \citet{guo-2017-cal} and use the development split of MC-train/Ext-train to optimize the parameters.\footnote{Since not all datasets in MC-test and Ext-test have a test split, we report the performance on the development split.}

\paragraph{LMs}
One clear trend of the past several years is that the parameter size and training data size of pre-trained models plays a significant role in the accuracy of models; pre-trained LMs such as BERT \cite{devlin-etal-2019-bert} tend to underperform more recently released larger LMs like Turing-NLG\footnote{\url{https://msturing.org/}} and GPT-3 \cite{brown-2020-gpt3}.
Thus, we use the largest publicly available LM, which at the time of this writing is \citet{raffel-2019-t5}'s T5 model.
The T5 model is a sequence-to-sequence model with both encoder and decoder using transformers \cite{vaswani-2017-attn}, and the largest version has 11 billion parameters, allowing it to realize state-of-the-art performance on tasks such as question answering and natural language understanding \cite{roberts-2020-pack,khashabi-2020-unifiedqa}.

Specifically, we use two varieties of this model.
The original \textbf{T5} model is a sequence-to-sequence model trained on a large corpus of web text, specifically trained on a denoising objective that generates missing tokens given inputs with some tokens masked out.
The \textbf{UnifiedQA} model, uses the initial T5 model and fine-tunes on a variety of QA datasets by converting multiple-choice, extractive QA formats into a unified sequence-to-sequence format, similar to the one that we show in \autoref{tab:example}.
We use the 3-billion versions in our main experiments in \autoref{sec:exp_main} (for efficiency purposes), but also report the performance of the largest 11-billion versions in ablation studies \autoref{sec:exp_abl}.

For comparison with LMs of different architectures trained on different datasets, we also report the performance of two other LMs in \autoref{sec:exp_lms}: the 0.4-billion BART model \cite{lewis-2020-bart} which is a sequence-to-sequence model and the 0.7-billion GPT-2 large model \cite{radford-2019-gpt2} which is a conventional language model.
We fine-tune them following the same recipe of UnifiedQA \cite{khashabi-2020-unifiedqa}.

\paragraph{Evaluation Metrics}
We use accuracy to measure the prediction performance of our methods, and ECE to measure the calibration performance.
Accuracy is computed as the ratio of question-answer pairs for which the correct answer has the highest probability among all the candidates in $\mathcal{I}(x)$.
ECE is computed using \autoref{eq:ece} by bucketing all candidate answers in $\mathcal{I}(x)$ based on confidence.
For MC-test and Ext-test which include multiple datasets, we compute accuracy and ECE on each dataset separately and average across them to avoid the metrics being dominated by large datasets.

\paragraph{Implementation Details}
We fine-tune UnifiedQA-3B with a batch size of 16 for 3k steps and UnifiedQA-11B with a batch size of 3 for 15k steps on a v3-8 TPU.
The maximal length of input and output are set to 512 and 128 respectively, following the setting of UnifiedQA \cite{khashabi-2020-unifiedqa}.
For extractive QA datasets, we use top $R=10$ first tokens and finally $K=5$ spans are used as candidates. 
For the paraphrasing-based method, we use the WMT-19 English-German and German-English transformer models to perform back translation \cite{ng-2019-fairwmt19}.
The beam size is set to 10 for both directions, which will yield $10 \times 10 = 100$ paraphrases in the end.
Since some paraphrases are duplicated, we count the frequency and use the top 5 unique paraphrases in our main experiments \autoref{sec:exp_main}.
We also report the performance of using different numbers of paraphrases in \autoref{sec:exp_abl}.
For the retrieval-based augmentation, we use the KILT toolkit \cite{petroni-2020-kilt} to retrieve the most relevant article from the Wikipeida dump, and append the first three sentences of the first paragraph of the retrieved article to the input.
For the feature-based decision trees model, we use XGBoost \cite{chen-2016-xgb} with logistic binary objective, max depth of 4, number of parallel trees of 5, and subsample ratio of 0.8.
We use \textbf{Temp.} to denote temperature-based scaling, \textbf{XGB} to denote feature-based decision trees, \textbf{Para.} to denote paraphrasing, \textbf{Aug.} to denote input augmentation, and \textbf{Combo} to denote the combination of Temp., Para., and Aug. in the experimental section.
We use the model with the best calibration performance in post-hoc calibration experiments.
For multiple-choice QA, we use the UnifiedQA model after margin-based fine-tuning.
For extractive QA, we use the original UnifiedQA model.

\subsection{Are LM-based QA Models Well Calibrated?}

As shown in \autoref{tab:finetune}, our baseline models (i.e., T5 and UnifiedQA) are strong, achieving state-of-the-art accuracy on a diverse range of QA datasets.
On the MT-test datasets, the UnifiedQA model even outperforms the largest version of GPT-3 with 175 billions parameters \cite{hendrycks-2020-mass}.
Despite the impressive performance, these models are not well calibrated, with ECE higher than 0.2 on the MT-test dataset.
We found that LMs tend to be over-confident about cases they do not know, as shown in the confidence distribution in the first row of \autoref{fig:dist} that most predictions have aggressive confidence being close to 0 or 1.
The UnifiedQA model assigns high confidence to the wrong answer for examples in \autoref{tab:example}, indicating that its confidence estimates are not trustworthy.

\subsection{Can LM-based QA Models be Calibrated?}\label{sec:exp_main}

We calibrate the UnifiedQA model using both fine-tuning-based methods and post-hoc methods and show their performance in \autoref{tab:finetune} and \autoref{tab:posthoc} respectively.

\begin{figure}[tb]
  \subfigure[T5]{
    \includegraphics[width=0.48\columnwidth, clip, keepaspectratio]{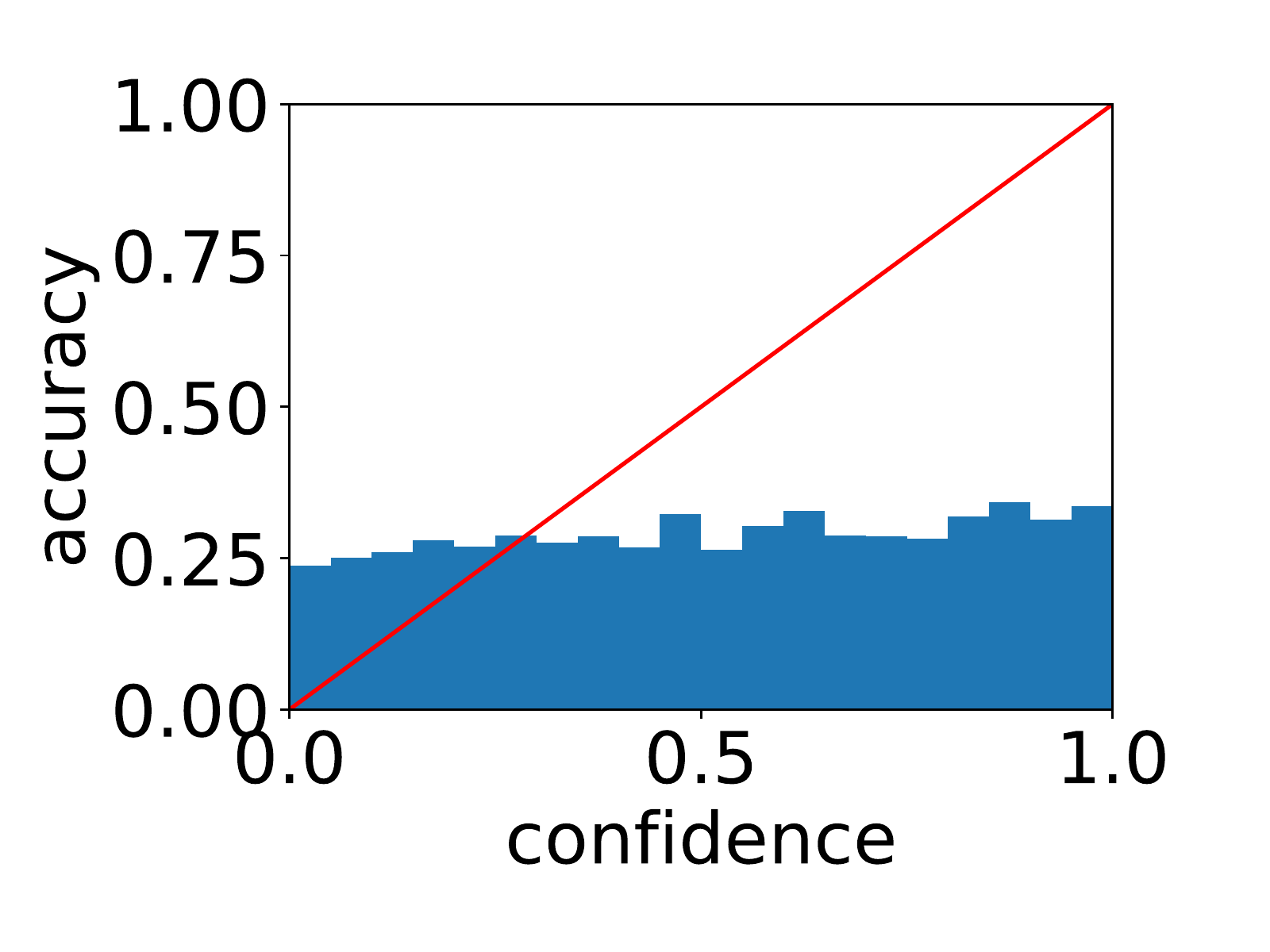}}
  \subfigure[UnifiedQA]{
    \includegraphics[width=0.48\columnwidth, clip, keepaspectratio]{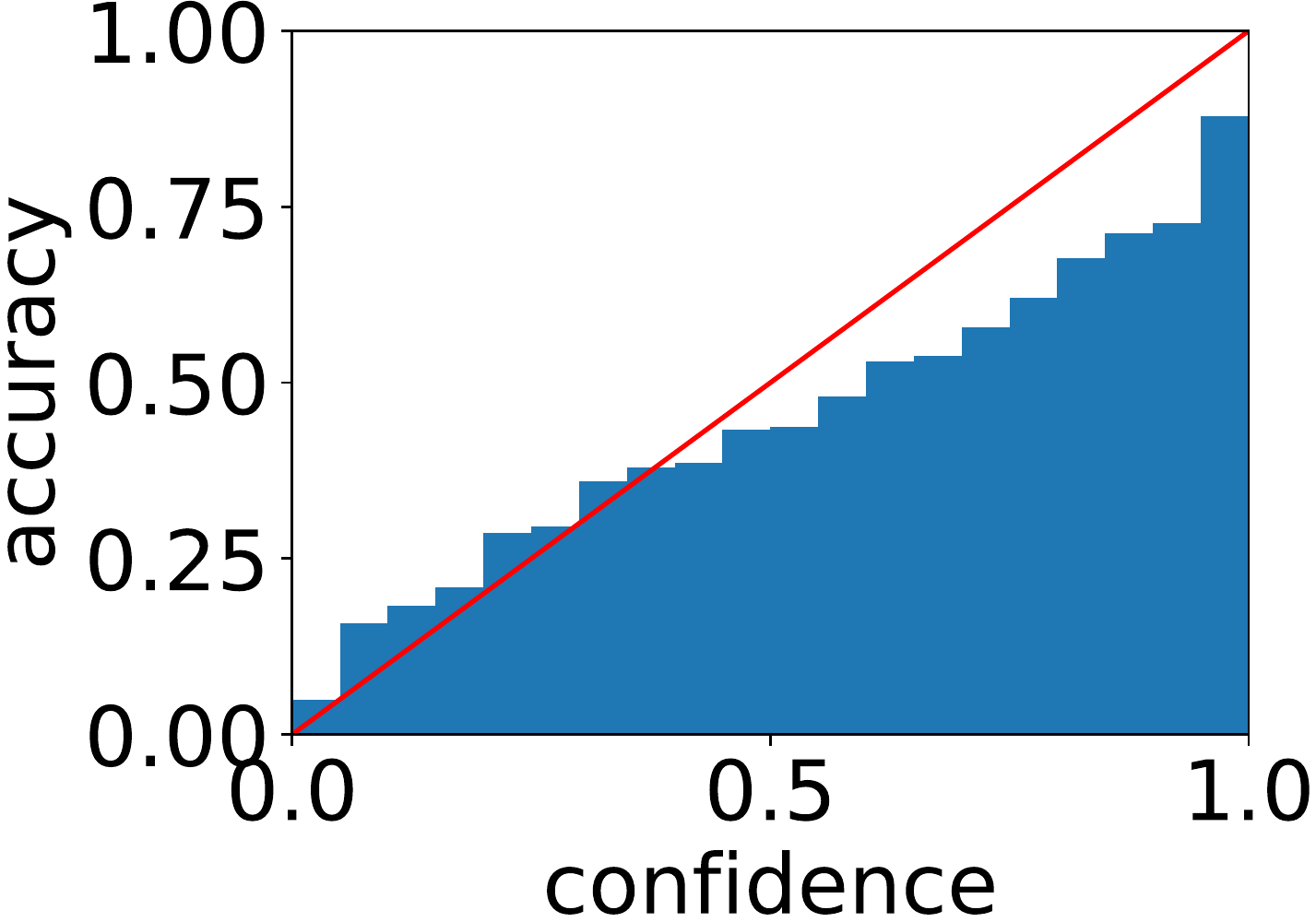}}
  \subfigure[UnifiedQA w/ Combo]{
    \includegraphics[width=0.48\columnwidth, clip, keepaspectratio]{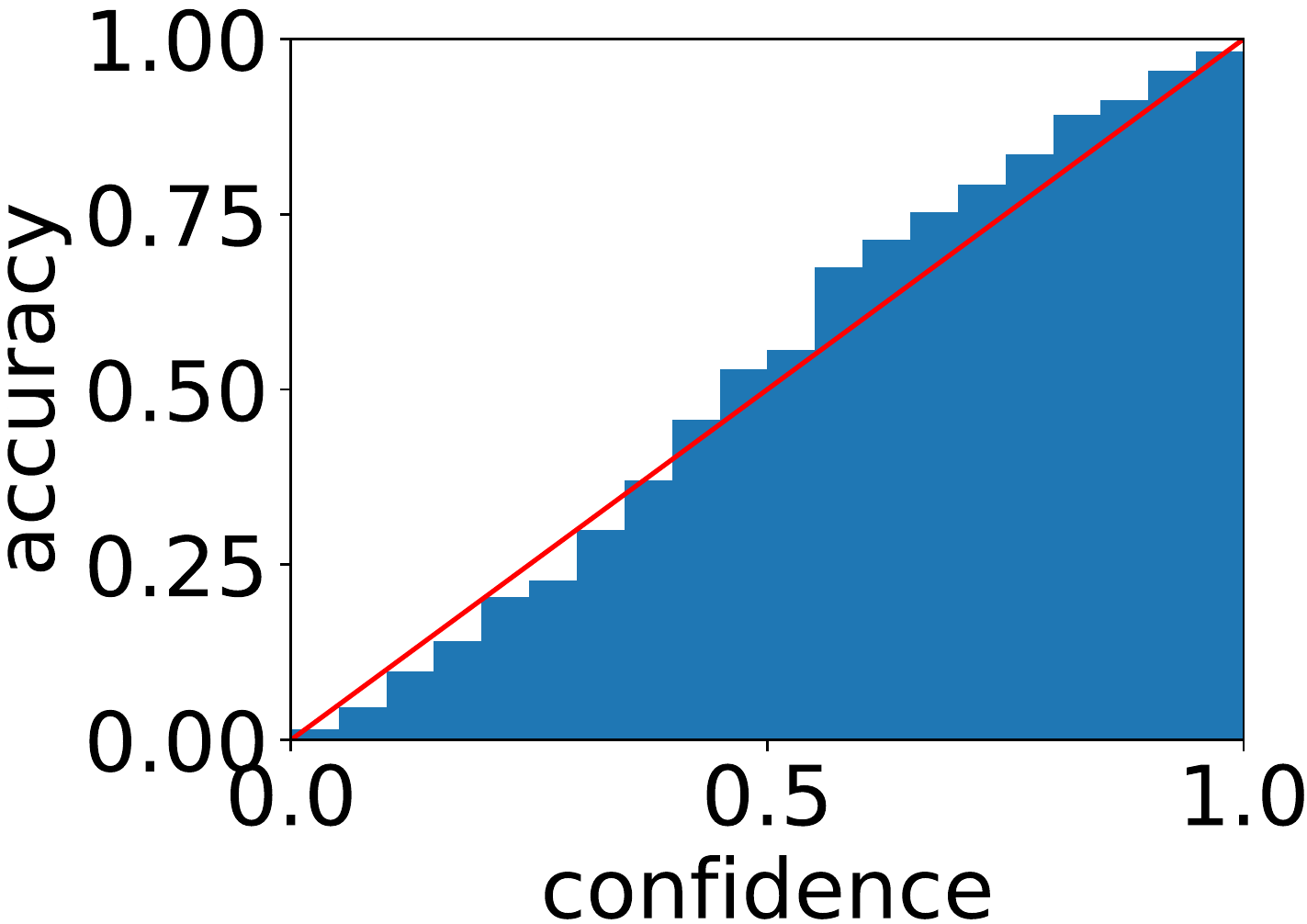}}
  \subfigure[UnifiedQA w/ Combo and oracle temperature]{
    \includegraphics[width=0.48\columnwidth, clip, keepaspectratio]{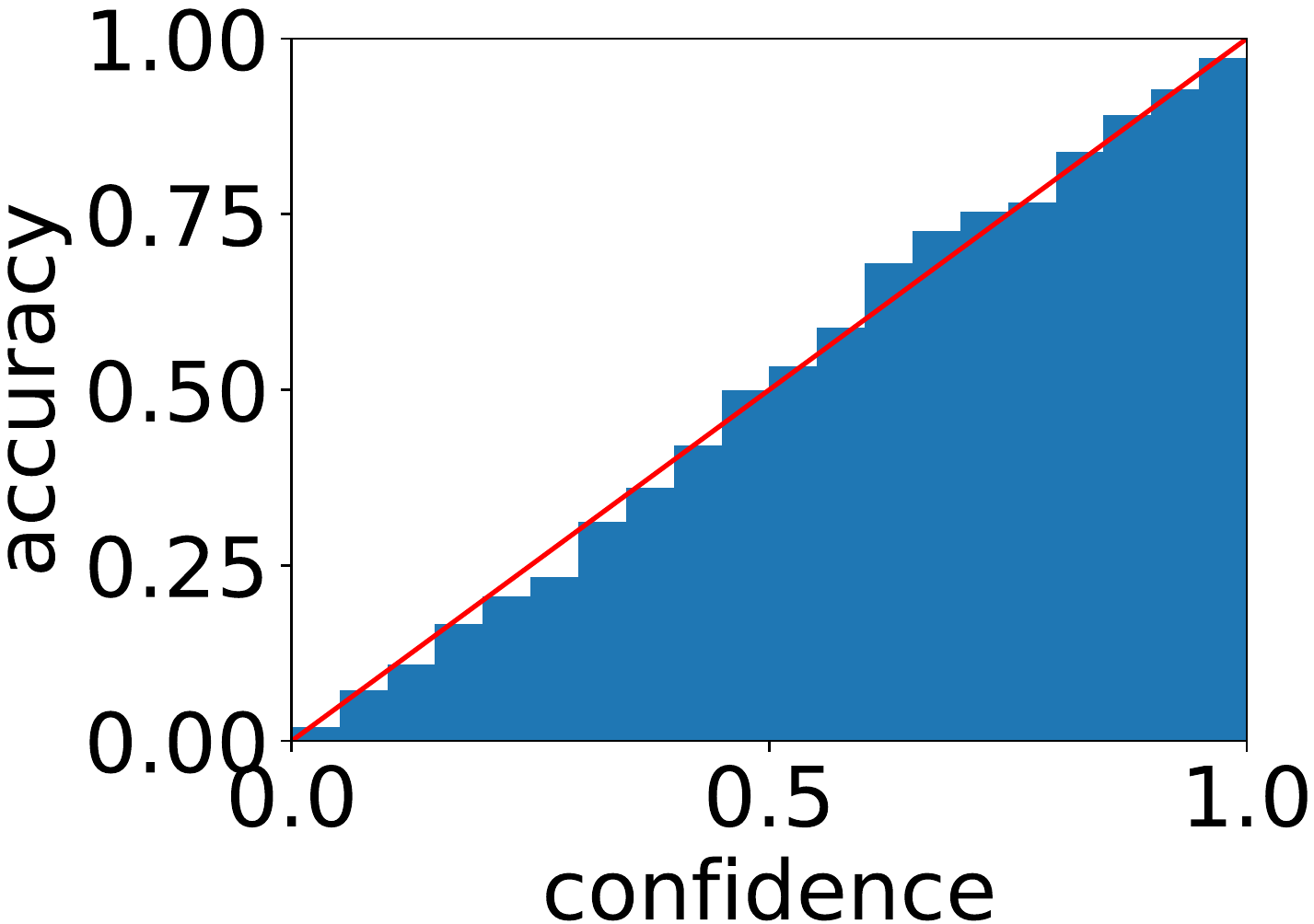}}
\caption{Reliability diagram of the T5 model (top-left), the original UnifiedQA model (top-right), the UnifiedQA model after calibration with Combo (bottom-left), and Combo with oracle temperature (bottom-right) on the MC-test datasets.
}
\label{fig:reliability}
\end{figure}

Overall, on multi-choice QA datasets (i.e., MC-test and MT-test), both fine-tuning-based methods and post-hoc methods can improve ECE while maintaining accuracy compared to the baseline UnifiedQA model.
The best-performing method (i.e., Combo), which combines margin-based fine-tuning, temperature-based scaling, paraphrasing, and input augmentation, improves ECE from 0.095 to 0.044 by over 53\%.
As shown in the reliability diagrams of the original UnifiedQA model (top-right) and the UnifiedQA model calibrated with Combo (bottom-left) in \autoref{fig:reliability}, calibration using our methods makes the confidence estimates of predictions better aligned with their correctness.
Comparing those two diagrams, an interesting observation is that our method seems to over-calibrate the LM, making over-estimated bars on the right-hand side of the top-right diagram (bars lower than the diagonal) under-estimated and vice versa.
This is probably caused by the temperature being too aggressive (i.e., too large), making the distribution too flat.
Note that the datasets used to learn the temperature (MC-train) and used in evaluation (MC-test) are different, which we hypothesize is the reason why the temperature is too aggressive.
We verify this by learning an oracle temperature on the evaluation datasets (MC-test).
The learned temperature indeed becomes smaller (1.35 $\rightarrow$ 1.13), and the reliability diagram (bottom-right in \autoref{fig:reliability}) is almost perfectly aligned.
This demonstrates the challenge of calibrating LMs across different domains.

However, on extractive QA datasets, the improvement brought by different calibration methods is smaller.
We hypothesize that this is because the candidate set $\mathcal{I}(X)$ generated by the span-based decoding method for extractive QA are harder to calibrate than the manually curated candidate answers for multiple-choice QA.
We compute the average entropy of the confidence of the UnifiedQA model over $\mathcal{I}(X)$ on both extractive QA (Ext-test) and multiple-choice QA datasets (MC-test), and found that Ext-test indeed has much higher entropy compared to MC-test (0.40 vs 0.13), which partially explains the difficulty of calibration on extractive QA datasets.

\begin{table}[tb]
\small
\centering
\begin{tabular}{@{}l@{\tinycol}c@{\tinycol}c|c@{\tinycol}c|c@{\tinycol}c@{}}
\toprule
\textbf{Method} & \multicolumn{2}{c|}{\textbf{MC-test}} & \multicolumn{2}{c|}{\textbf{MT-test}} & \multicolumn{2}{c}{\textbf{Ext-test}} \\
 & \textbf{ACC} & \textbf{ECE} & \textbf{ACC} & \textbf{ECE} & \textbf{ACC} & \textbf{ECE} \\
\midrule
T5 & 0.313 & 0.231 & 0.268 & 0.248 & 0.191 & 0.166 \\  
UnifiedQA & 0.769 & 0.095 & 0.437 & 0.222 & 0.401 & 0.114 \\
\quad + softmax & 0.767 & 0.065 & 0.433 & 0.161 & 0.394	& \textbf{0.110} \\
\quad + margin & 0.769 & \textbf{0.057} & 0.431 & \textbf{0.144} & 0.391 & 0.112 \\ 
\bottomrule
\end{tabular}
\caption{Performance of different fine-tuning methods.}
\label{tab:finetune}
\end{table}

\begin{table}[tb]
\small
\centering
\begin{tabular}{@{}l@{\tinycol}c@{\tinycol}c|c@{\tinycol}c|c@{\tinycol}c@{}}
\toprule
\textbf{Method} & \multicolumn{2}{c|}{\textbf{MC-test}} & \multicolumn{2}{c|}{\textbf{MT-test}} & \multicolumn{2}{c}{\textbf{Ext-test}} \\
 & \textbf{ACC} & \textbf{ECE} & \textbf{ACC} & \textbf{ECE} & \textbf{ACC} & \textbf{ECE} \\
\midrule
Baseline & 0.769 & 0.057 & 0.431 & 0.144 & 0.401 & 0.114 \\ 
\midrule
\quad + Temp. & 0.769 & 0.049 & 0.431 & \textbf{0.075} & 0.401 & 0.107 \\
\quad + XGB & 0.771 & 0.055 & 0.431 & 0.088 & 0.402 & \textbf{0.103} \\
\quad + Para. & 0.767 & 0.051 & 0.429 & 0.122 & 0.393 & 0.114 \\
\quad + Aug. & 0.744 & 0.051 & 0.432 & 0.130 & 0.408 & 0.110 \\
\midrule
\quad + Combo & 0.748 & \textbf{0.044} & 0.431 & 0.079 & 0.398 & 0.104 \\
\bottomrule
\end{tabular}
\caption{Performance of different post-hoc methods using the UnifiedQA model after margin-based fine-tuning or the original UnifiedQA model as the baseline model. ``+Combo'' denotes the method using both Temp., Para., and Aug.}
\label{tab:posthoc}
\end{table}

\subsection{Analysis of Individual Calibration Methods}
In this section, we discuss each method in detail and analyze why they can improve calibration performance.

\paragraph{Objective Function Matters.}
The original UnifiedQA model is fine-tuned based on MLE, which maximizes the probability of the gold answer given the question.
Both softmax-based and margin-based fine-tuning, which explicitly compare and adjust the probability of candidate answers, can further improve ECE on multiple-choice datasets.
We argue that the softmax-based and margin-based objective functions are better suited for questions with potential candidates.

\begin{figure}[tb]
  \subfigure[T5]{
    \includegraphics[width=0.48\columnwidth, clip, keepaspectratio]{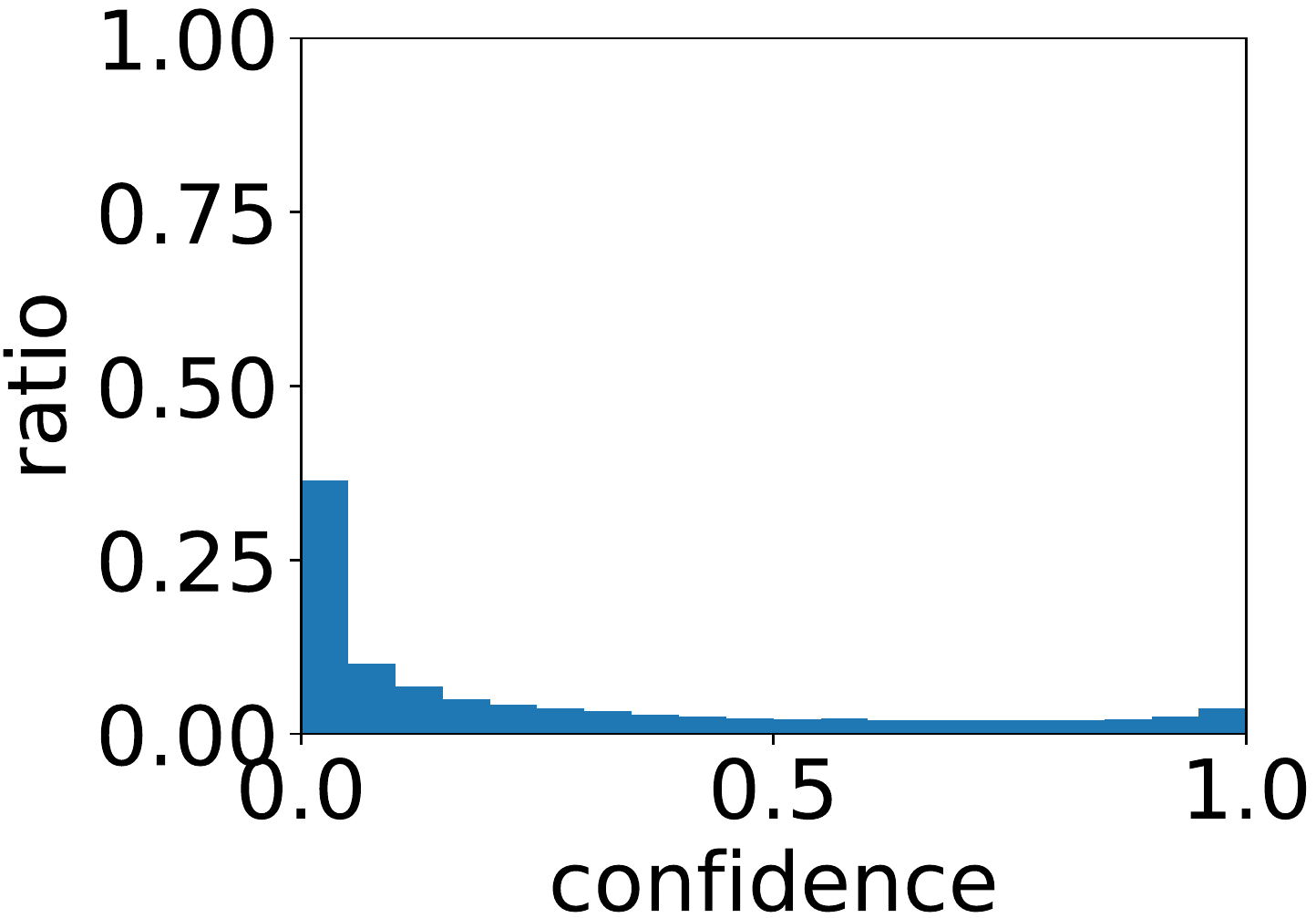}}
  \subfigure[UnifiedQA]{
    \includegraphics[width=0.48\columnwidth, clip, keepaspectratio]{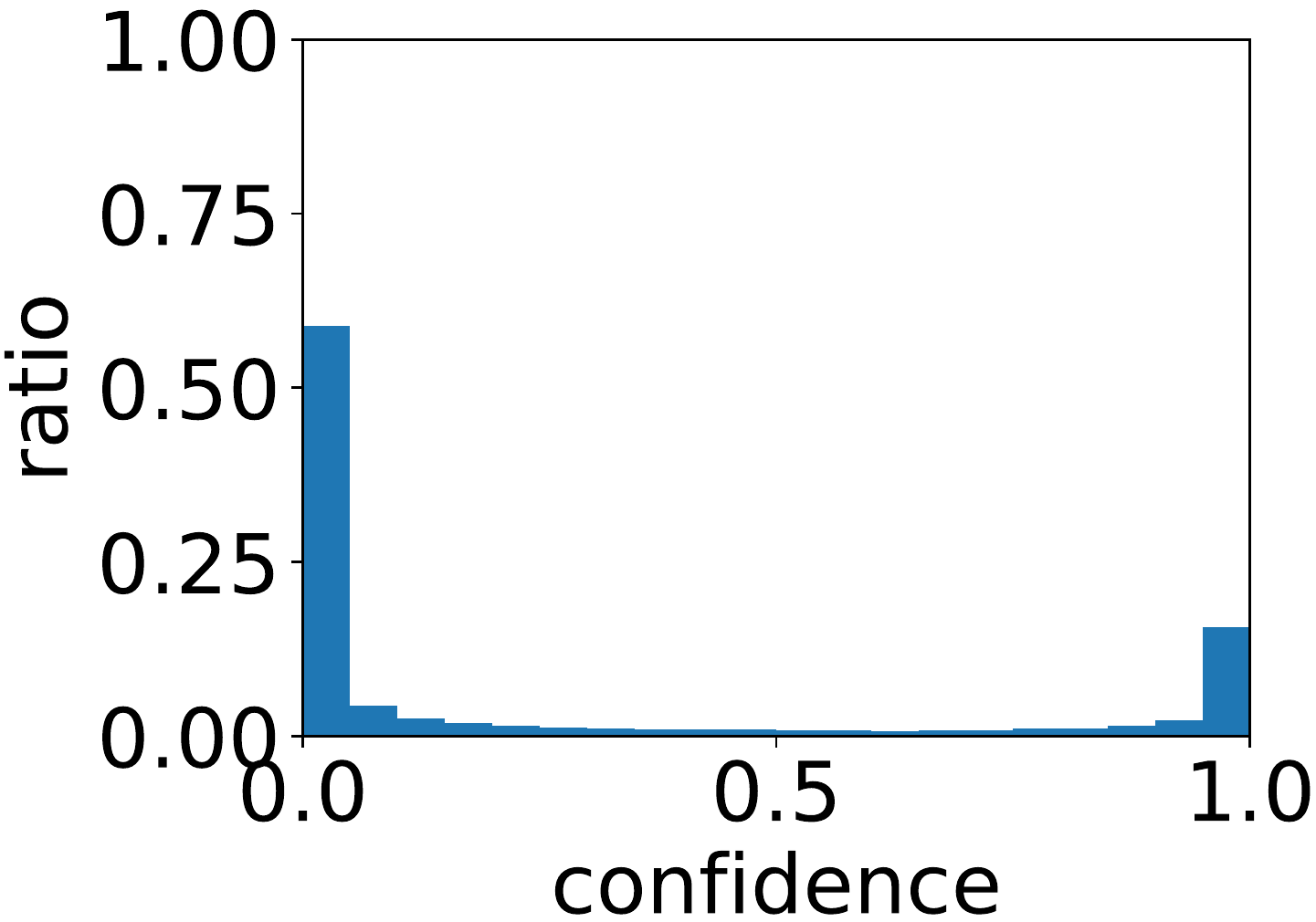}}
  \subfigure[UnifiedQA w/ Temp.]{
    \includegraphics[width=0.48\columnwidth, clip, keepaspectratio]{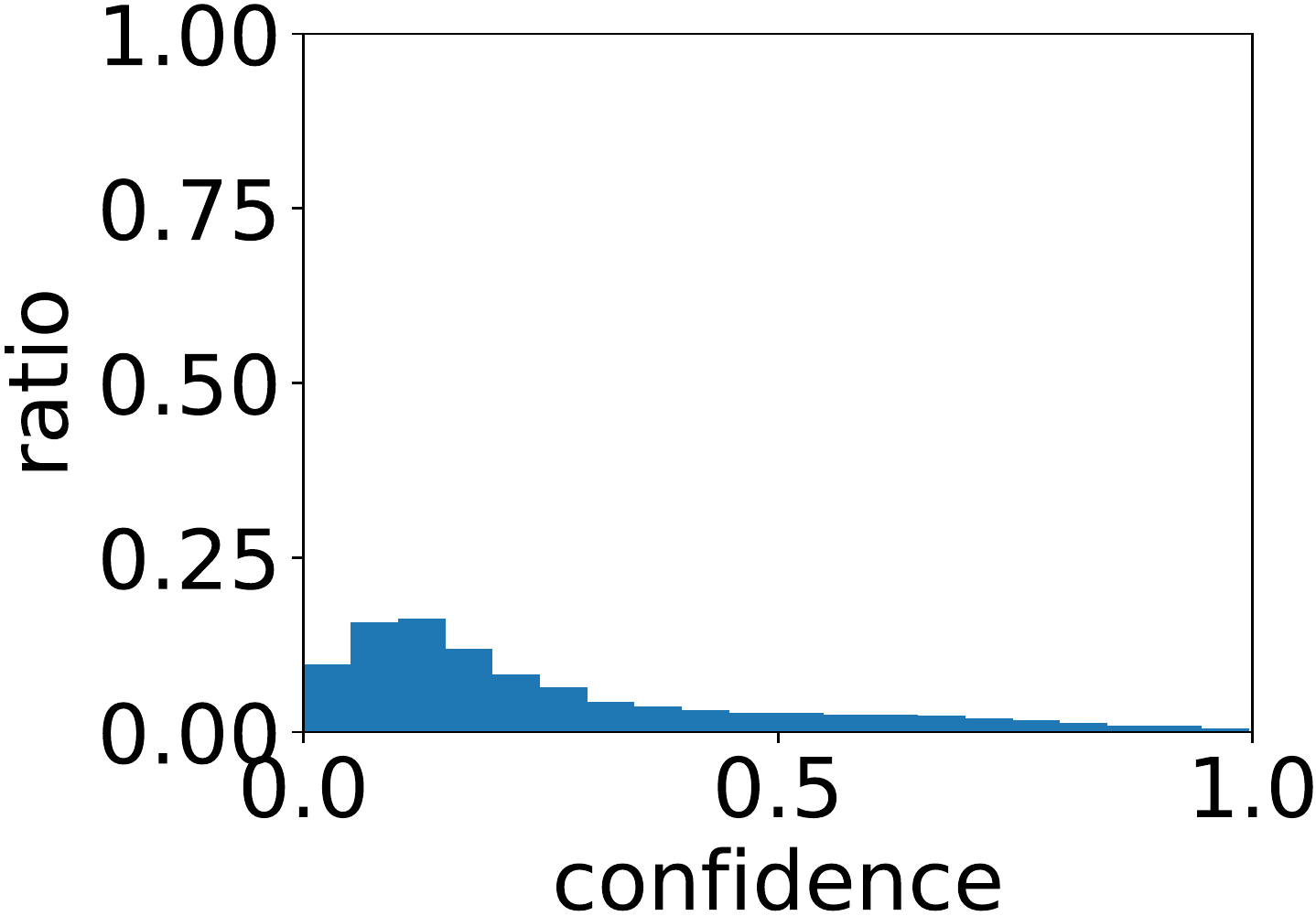}}
  \subfigure[UnifiedQA w/ XGB]{
    \includegraphics[width=0.48\columnwidth, clip, keepaspectratio]{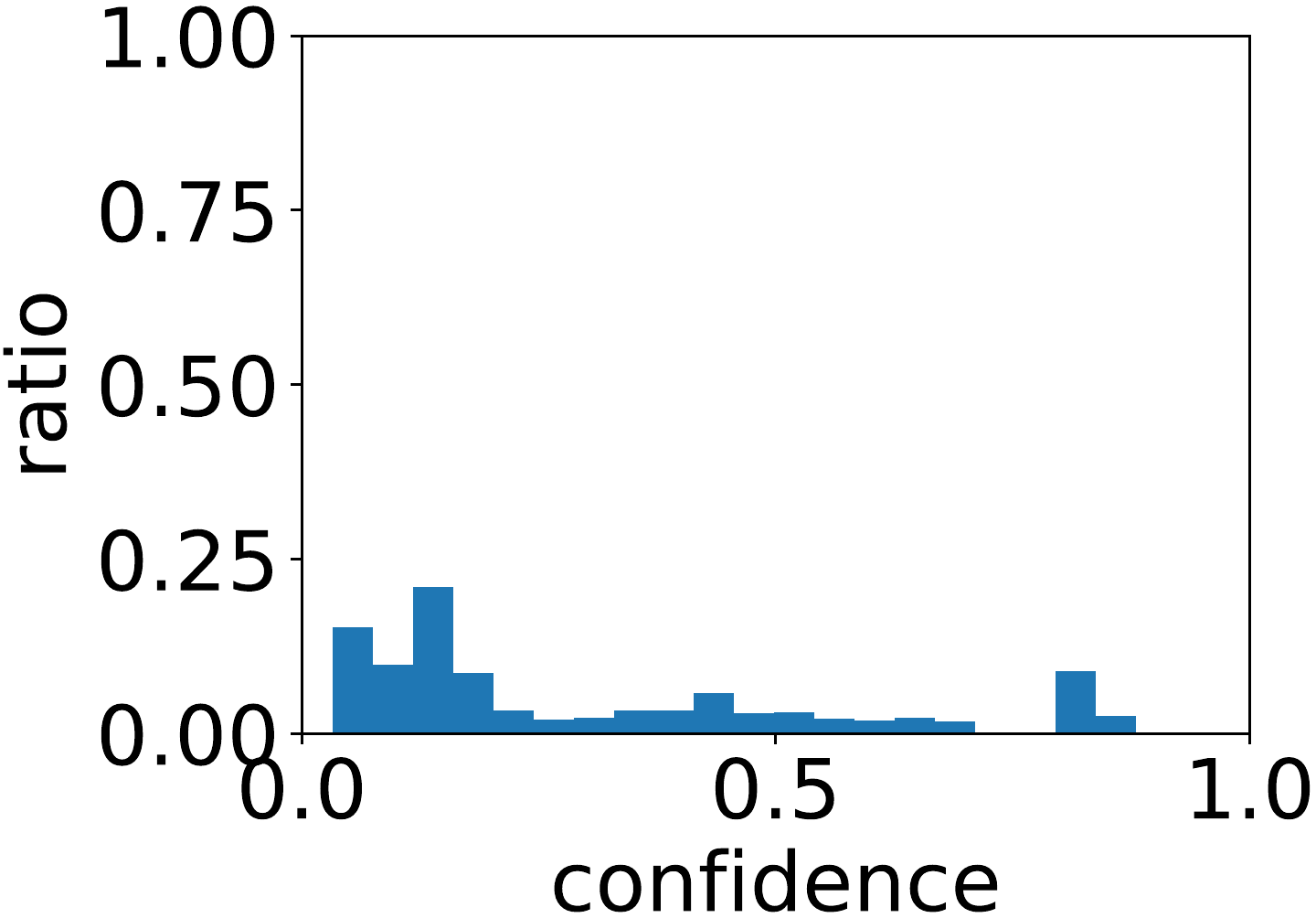}}
\caption{The ratio of predictions with respect to confidence of the T5 model (top-left), the UnifiedQA model (top-right), the UnifiedQA model after temperature-based calibration (bottom-left), and the UnifiedQA model after feature-based calibration (bottom-right) on the MC-test datasets.}
\label{fig:dist}
\end{figure}

\paragraph{Post-processing Confidence is Effective Universally.}
Post-processing the raw confidence either solely based on confidence or other features is effective across all datasets, which is consistent with the conclusion on other tasks such as structured prediction and natural language inference \cite{jagannatha-2020-structcal, desai-2020-transcal}.
We demonstrate the histogram of confidence before and after applying temperature-based scaling or feature-based decision trees in \autoref{fig:dist}.
LMs tend to be over-confident, with most predictions having either extremely high or low confidence.
Both methods can successfully re-scale the confidence to reasonable ranges, thus improving the calibration performance.

\paragraph{Paraphrasing Answers and Input Augmentation can Improve Confidence Estimation.}
The improvement brought by using paraphrasing is significant on multiple-choice datasets, demonstrating that using diverse expressions can indeed improve confidence estimation.
To better understand under what circumstances paraphrasing works, we group candidate answers into two categories: the first group includes candidate answers that become better calibrated using paraphrases; the second group includes candidate answers whose confidence remains the same using paraphrases.
We say that a candidate becomes better calibrated if its confidence increases/decreases by 20\% if it is a correct or incorrect answer respectively.
We found that the average length of questions for better calibrated candidates (187) is much shorter than that of candidates without improvement (320), indicating that paraphrasing is useful mainly for short questions.
We also compute the diversity of word usage in paraphrases using the number of unique words divided by the total length of paraphrases.
We found that better calibrated candidates have slightly higher diversity (0.35 vs 0.32), which is consistent with our intuition.
Retrieval-based augmentation can also improve calibration performance on multiple-choice datasets, which is probably because the retrieved documents can provide extra evidence about the question, making LMs more robust at confidence estimation.

\paragraph{Calibration Methods are Complementary.}
By combining margin-based fine-tuning, temperature-based scaling, paraphrasing, and input augmentation, we achieve the best ECE on MC-test, demonstrating that these calibration methods are complementary to each other.

\subsection{Ablation Study}\label{sec:exp_abl}

In this section, we perform an ablation study to examine different aspects of LM calibration, including calibration performance of different LMs, across LMs with different sizes, using different numbers of paraphrases, and across datasets with potential domain shift.

\paragraph{Performance of Different LMs.}\label{sec:exp_lms}

\begin{table}[tb]
\small
\centering
\begin{tabular}{lc@{\tinycol}c|c@{\tinycol}c}
\toprule
\textbf{Method} & \multicolumn{2}{c|}{\textbf{BART}} & \multicolumn{2}{c}{\textbf{GPT-2 large}} \\
 & \textbf{ACC} & \textbf{ECE} & \textbf{ACC} & \textbf{ECE} \\
\midrule
Original & 0.295 & 0.225 & 0.272 & 0.244 \\
+ UnifiedQA & 0.662 & 0.166 & 0.414 & 0.243 \\
\quad + softmax & 0.658 & 0.097 & 0.434 & 0.177 \\
\quad + margin & 0.632 & 0.090 & 0.450 & 0.123 \\
\midrule
\quad\quad + Temp. & 0.632 & \textbf{0.064} & 0.450 & \textbf{0.067} \\
\quad\quad + XGB & 0.624 & 0.090 & 0.440 & 0.080 \\
\quad\quad + Para. & 0.624 & 0.084 & 0.436 & 0.104 \\
\quad\quad + Aug. & 0.600 & 0.089 & 0.441 & 0.126 \\
\midrule
\quad\quad + Combo & 0.591 & 0.065 & 0.429 & 0.069 \\
\bottomrule
\end{tabular}
\caption{Performance of different LMs on the MC-test dataset. ``Original'' indicates the original language model, and ``+ UnifiedQA'' indicates fine-tuning following the recipe of UnifiedQA.}
\label{tab:lms}
\end{table}

We report the performance of two other LMs in \autoref{tab:lms}. Both the BART and GPT-2 models are smaller than T5, thus the overall accuracy and calibration performance are lower than that of T5.
Both fine-tuning and post-hoc calibration methods can improve ECE, indicating that our methods are applicable to LMs trained with different datasets and architectures.

\paragraph{Performance of LMs with Different Sizes.}
We conduct experiments using the largest version (i.e., 11B) of the T5 and UnifiedQA model to analyze how calibration performance varies with respect to the size of the LM in \autoref{tab:11b}.
We found that larger LMs usually achieve both higher accuracy and better calibration performance, which is contradictory to the observation in image classification \cite{guo-2017-cal} where larger models such as ResNet \cite{he-2016-resnet} are no longer well calibrated compared to smaller models like LeNet \cite{lecun-1998-lenet}.
Given the fact the size of both the pre-training corpus and LMs are extremely larger compared to previous practice, we might have completely different observations with respect to confidence estimation.
Unlike ResNet trained on CIFAR-100, the training of LMs is not bottlenecked by the dataset, and larger LMs have a stronger capacity to model text distribution and memorize facts, which leads to better calibration performance overall \cite{kaplan-2020-scalinglm}.
Overall, our methods can improve ECE from 0.067 to 0.032 using the 11B UnifiedQA model on the MC-test dataset, and from 0.175 to 0.085 on the MT-test dataset.
However, compared to the 3B version, improvement brought by post-hoc calibration methods is smaller, which is probably because the 11B version is better optimized and more knowledgeable.

\begin{table}[tb]
\small
\centering
\begin{tabular}{lc@{\tinycol}c|c@{\tinycol}c}
\toprule
\textbf{Method} & \multicolumn{2}{c|}{\textbf{MC-test}} & \multicolumn{2}{c}{\textbf{MT-test}} \\
 & \textbf{ACC} & \textbf{ECE} & \textbf{ACC} & \textbf{ECE} \\
\midrule
T5 & 0.359 & 0.206 & 0.274 & 0.235 \\
UnifiedQA & 0.816 & 0.067 & 0.479 & 0.175 \\
\quad + softmax & 0.823 & 0.041 & 0.488 & 0.129 \\
\quad + margin & 0.819 & 0.034 & 0.485 & 0.107 \\
\midrule
\quad\quad + Temp. & 0.819 & 0.036 & 0.485 & 0.098 \\
\quad\quad + XGB & 0.818 & 0.065 & 0.486 & 0.108 \\
\quad\quad + Para. & 0.820 & 0.035 & 0.484 & 0.092 \\
\quad\quad + Aug. & 0.812 & \textbf{0.031} & 0.493 & 0.090 \\
\midrule
\quad\quad + Combo & 0.807 & 0.032 & 0.494 & \textbf{0.085} \\
\bottomrule
\end{tabular}
\caption{Performance of the 11B LMs.}
\label{tab:11b}
\end{table}

\paragraph{Performance using Different Numbers of Paraphrases.}

\begin{figure}[tb]
\centering
\includegraphics[width=1.0\columnwidth, clip, keepaspectratio]{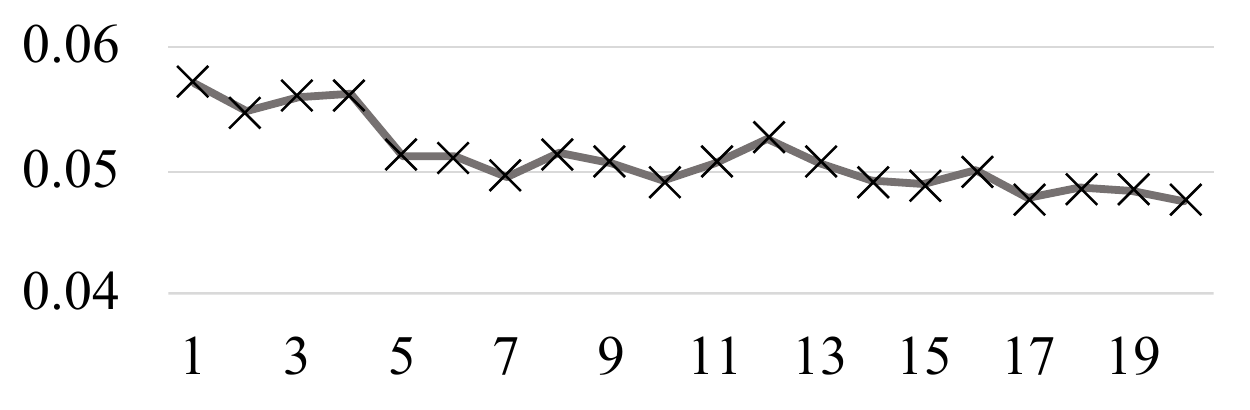}
\caption{ECE of the UnifiedQA model using different numbers of paraphrases on the MC-test datasets.}
\label{fig:para}
\end{figure}

In \autoref{fig:para}, we experiment with different numbers of paraphrases using the UnifiedQA model on MC-test datasets.
The overall trend is that the more paraphrases we use, the better calibrated the LM, demonstrating that using different variations to express the candidate answer can improve confidence estimation.
The improvements using more than 10 paraphrases are subtle, so 5-10 paraphrases may represent a good trade-off between computational cost and performance in practical settings.

\paragraph{Performance on Training and Evaluation Datasets.}
As introduced in the experimental section, we perform calibration on the MC-train dataset and evaluate the final performance on the MC-test dataset to study whether our calibration methods can generalize to out-of-domain dataset.
We compare the performance on the training dataset and the evaluation dataset in \autoref{tab:inout}.
We found that on both datasets, each individual method can improve ECE, indicating that our method can generalize to out-of-domain datasets.
Note that the improvement on the training dataset (0.133 $\rightarrow$ 0.042) is larger than on improvement on the evaluation dataset (0.095 $\rightarrow$ 0.044), which is probably caused by the domain shift between the two datasets.

\begin{table}[tb]
\small
\centering
\begin{tabular}{lc@{\tinycol}c|c@{\tinycol}c}
\toprule
\textbf{Method} & \multicolumn{2}{c|}{\textbf{MC-train}} & \multicolumn{2}{c}{\textbf{MC-test}} \\
 & \textbf{ACC} & \textbf{ECE} & \textbf{ACC} & \textbf{ECE} \\
\midrule
T5 & 0.334 & 0.228 & 0.313 & 0.231 \\
UnifiedQA & 0.727 & 0.133 & 0.769 & 0.095 \\
\quad + softmax & 0.735 & 0.084 & 0.767 & 0.065 \\
\quad + margin & 0.737 & 0.069 & 0.769 & 0.057 \\
\midrule
\quad\quad + Temp. & 0.737 & 0.051 & 0.769 & 0.049 \\
\quad\quad + XGB & 0.737 & 0.074 & 0.771 & 0.055 \\
\quad\quad + Para. & 0.742 & 0.053 & 0.767 & 0.051 \\
\quad\quad + Aug. & 0.721 & 0.059 & 0.744 & 0.051 \\
\midrule
\quad\quad + Combo & 0.722 & \textbf{0.042} & 0.748 & \textbf{0.044} \\
\bottomrule
\end{tabular}
\caption{Performance comparison between training and evaluation datasets.}
\label{tab:inout}
\end{table}

\section{Related Work}
\paragraph{Calibration}
Calibration is a well-studied topic in other tasks such as medical diagnosis \cite{jiang-2010-medcal} and image recognition \cite{guo-2017-cal,lee-2018-cal}.
Previous works in NLP have examined calibration in structured prediction problems such as part-of-speech tagging and named entity recognition \cite{jagannatha-2020-structcal}, natural language understanding tasks such as natural language inference, paraphrase detection, extractive question answering, and text classification \cite{desai-2020-transcal,kamath-2020-qacal,kong-2020-lminout}.
In contrast, we focus on calibrating LMs themselves by treating them as natural language generators that predict the next words given a particular input.

\paragraph{LM probing}
Previous works probe pre-trained LMs with respect to syntactic and semantic properties \cite{hewitt-2019-structprob,tenney-2019-bertpipe}, factual knowledge \cite{petroni-etal-2019-language,poerner-2019-ebert,jiang-2019-lpaqa}, commonsense knowledge \cite{trinh-2018-commonsense,kocijan-2019-wsc}, and other properties \cite{talmor-2019-olmpics}.
These works usually focus on what LMs know, while in this paper we also consider the cases when LMs do not know the answer with confidence.

\section{Conclusion}
In this paper, we examine the problem of calibration in LMs used for QA tasks.
We first note that despite the impressive performance state-of-the-art LM-based QA models tend to be poorly calibrated in their probability estimates.
To alleviate this problem, we attempted several methods to either fine-tune the LMs, or adjust the confidence by post-processing raw probabilities, augmenting inputs, or paraphrasing candidate answers.
Experimental results demonstrate the effectiveness of these methods.
Further analysis reveals the challenges of this problem, shedding light on future work on calibrating LMs.

Some future directions could be developing calibration methods for LMs on a more fine-grained level than simply holistic calibration across the entire dataset.
For example, there has been significant interest in how models perform across diverse subsets of the entire training data \citep{hashimoto-2018-fair} and how they reflect dataset biases \citep{rudinger-2018-gender}, and the interaction of model confidence with these phenomena is of significant interest.
It is also interesting to investigate the effect of calibration on users or downstream tasks.
For instance, providing users with model confidences can influence downstream decisions \citep{zhang-2020-trust}, and users may want to adjust required confidence thresholds on critical domains (e.g., health, safety, medicine).
All of these are interesting paths of inquiry for future research.

\section*{Acknowledgements}
This work was supported in part by a gift from Bosch research.
The authors thank the Google Cloud and TensorFlow Research Cloud for computation credits that aided in the execution of this research.

\bibliography{tacl2018}
\bibliographystyle{acl_natbib}

\end{document}